\documentclass{ifacconf}

\usepackage{graphicx}      
\usepackage{natbib}        
\usepackage{siunitx}
\usepackage{tikz}
\usepackage{pgf}
\usepackage{amsfonts}
\usepackage[acronym, toc]{glossaries}
\usepackage{mathtools}
\usepackage{bm}
\usepackage{siunitx}
\usepackage{subcaption}
\usepackage{multirow}
\usepackage{makecell}
\usepackage{xcolor}
\usepackage{caption}

\makeglossaries
\newglossaryentry{opencv}{
    name=OpenCV,
    description={The Open Computer Vision Library (OpenCV) is an open source computer vision library, mainly for computer vision and machine vision applications. It is an optimized implementation of a number of state-of-the-art computer vision algorithms, including image processing, computer vision, and computer graphics.},
}

\newglossaryentry{target}{
    name=target,
    description={The target is a non-cooperative target, tambling in the environment with constant velocities.},
}

\newglossaryentry{chaser}{
    name=chaser,
    description={The chaser, in the scope of this work, is the \textit{controlled} robotic system that alone, or part of a swarm, attempts to dock the \gls{target}. It is capable of 6-\acrshort{dof} motion in the environment and is equiped with various sensor hardware.},
}

\newglossaryentry{vicon}{
    name=Vicon,
    description={Vicon is a motion capture system that track objects in a laboratory environment. It uses multiple cameras mounted around a room that recognize special markers on the tracked objects. Vicon is produced by \href{https://www.vicon.com/}{Vicon Motion Systems Ltd}.},
}

\newglossaryentry{lidar}{
    name=LiDAR,
    description={LiDAR is an acronym for Light Detection and Ranging or Light Imaging, Detection and Ranging. It is a method of range measurement that uses a laser beam to measure time-of-flight between the sensor and an object or surface. LiDAR sensors can be single-beam, 2D or 3D, providing different formats of data such as distance, depth maps or point-clouds.}
}
\newacronym{slam}{SLAM}{Simultanous Localization and Mapping}
\newacronym{vo}{VO}{Visual Odometry}
\newacronym{pnp}{PnP}{Perspective-n-Point}
\newacronym{cnn}{CNN}{Convolutinal Neural Network}
\newacronym{kf}{KF}{Kalman Filter}
\newacronym{ekf}{EKF}{Extended Kalman Filter}
\newacronym{ransac}{RANSAC}{RANdom SAmple Consensus}
\newacronym{gps}{GPS}{Global Positioning System}
\newacronym{icp}{ICP}{Iterative Closest Point}
\newacronym{ros}{ROS}{Robot Operating System}
\newacronym{os}{OS}{Operating System}
\newacronym{dof}{DOF}{Degrees of Freedom}
\newacronym{api}{API}{Application Programming Interface}
\newacronym{rtos}{RTOS}{Real-Time Operating System}
\newacronym{lts}{LTS}{Long-Term Support}
\newacronym{qos}{QoS}{Quality of Service}
\newacronym{urdf}{URDF}{Unified Robotics Description Format}
\newacronym{xml}{XML}{Extensible Markup Language}
\newacronym{tf}{tf}{Transfer Function}
\newacronym{sdf}{SDF}{Simulation Description Format}
\newacronym{xacro}{XACRO}{XML Macro Language}
\newacronym{tsl}{TSL}{Thruster Selection Logic}
\newacronym{pwm}{PWM}{Pulse Width Modulation}
\newacronym{dsp}{DSP}{Digital Signal Processing}
\newacronym{mpc}{MPC}{Model Predictive Control}
\newacronym{ar}{AR}{Augmented Reality}
\newacronym{vr}{VR}{Virtual Reality}
\newacronym{epnp}{EPnP}{Efficient PnP}
\newacronym{uav}{UAV}{Unmanned Aerial Vehicles}
\newacronym{ode}{ODEs}{Ordinary Differential Equations}
\newacronym{mimo}{MIMO}{Multiple Input Multiple Output}
\newacronym{oos}{OOS}{On-Orbit Servicing}
\newacronym{iss}{ISS}{International Space Station}
\newacronym{gnss}{GNSS}{Global Navigation Satellite System}
\newacronym{hubble}{HST}{Hubble Space Telescope}
\newacronym{jwst}{JWST}{James Webb Space Telescope}
\newacronym{nasa}{NASA}{National Aeronautics and Space Administration}
\newacronym{esa}{ESA}{European Space Agency}
\newacronym{csa}{CSA}{Canadian Space Agency}
\newacronym{sqp}{SQP}{Sequencial Quadratic Programming}
\newacronym{qp}{QP}{Quadratic Programming}
\newacronym{open}{OpEn}{Optimization Engine}
\newacronym{nlp}{NLP}{Nonlinear Programming}

\graphicspath{{images/}}

\tikzstyle{rect}=[fill=none, draw=black, shape=rectangle, align=center, rounded corners=2pt, thick]
\tikzstyle{diamond}=[fill=none, draw=black, shape=diamond, thick, rounded corners=1pt, align=center]
\tikzstyle{connection}=[fill=black, draw=black, shape=circle, scale=.4]
\tikzstyle{rect_warn}=[fill={rgb,255: red,216; green,216; blue,216}, draw=black, shape=rectangle, align=center, rounded corners=2pt, thick]
\tikzstyle{circle}=[fill=none, draw=black, shape=circle, thick, align=center]
\tikzstyle{circlecross}=[minimum size=15pt, fill=none, draw=black, shape=circle, cross, thick]
\tikzstyle{crossout}=[fill=none, draw=black, shape=cross out, align=center]

\tikzstyle{arrow}=[thick, draw=black, ->]
\tikzstyle{category_wrapper}=[-, very thick]
\tikzstyle{arrow part}=[-, thick, draw=black]
\tikzstyle{double arrow}=[double, thick, draw=black, ->]
\tikzstyle{inter}=[double, thick, draw=black, <->]
\tikzstyle{result}=[very thick, draw=black, ->]

\usetikzlibrary{backgrounds}
\usetikzlibrary{arrows}
\usetikzlibrary{shapes,shapes.geometric,shapes.misc}

\tikzstyle{tikzfig}=[baseline=-0.25em,scale=0.5]

\pgfkeys{/tikz/tikzit fill/.initial=0}
\pgfkeys{/tikz/tikzit draw/.initial=0}
\pgfkeys{/tikz/tikzit shape/.initial=0}
\pgfkeys{/tikz/tikzit category/.initial=0}

\pgfdeclarelayer{edgelayer}
\pgfdeclarelayer{nodelayer}
\pgfsetlayers{background,edgelayer,nodelayer,main}

\tikzstyle{none}=[inner sep=0mm]

\newcommand{\tikzfig}[1]{%
{\tikzstyle{every picture}=[tikzfig]
\IfFileExists{#1.tikz}
  {\resizebox{.95\linewidth}{!}{\input{#1.tikz}}}
  {%
    \IfFileExists{./images/#1.tikz}
      {\resizebox{.95\linewidth}{!}{\input{./images/#1.tikz}}}
      {\tikz[baseline=-0.5em]{\node[draw=red,font=\color{red},fill=red!10!white] {\textit{#1}};}}%
  }}%
}

\newcommand{\insertfig}[2]{%
  \begin{figure}[h]
    \centering
    \input{#1.tikz}
    \caption{#2}
    \label{fig:#1}
  \end{figure}
}

\tikzstyle{every loop}=[]

\DeclareMathOperator{\dist}{dist}

\begin{document}
\begin{frontmatter}

\title{
    Vision Based Docking of Multiple Satellites with an Uncooperative Target \thanksref{footnoteinfo}
} 

\thanks[footnoteinfo]{© 2023 Fragiskos Fourlas, Vignesh Kottayam Viswanathan, Sumeet Satpute, and George Nikolakopoulos. This work has been accepted to IFAC for publication under a Creative Commons Licence CC-BY-NC-ND}

\author{Fragiskos Fourlas}, 
\author{Vignesh Kottayam Viswanathan},
\author{Sumeet Satpute}, and
\author{George Nikolakopoulos}


\address{Robotics and AI Team\\ Lule\aa\,\,University of Technology, Lule\aa\, Sweden\\E-mail: \{frafou; vigkot; sumsat; geonik \}@ltu.se}

\begin{abstract}                
With the ever growing number of space debris in orbit, the need to prevent further space population is becoming more and more apparent. Refueling, servicing, inspection and deorbiting of spacecraft are some example missions that require precise navigation and docking in space. Having multiple, collaborating robots handling these tasks can greatly increase the efficiency of the mission in terms of time and cost. This article will introduce a modern and efficient control architecture for satellites on collaborative docking missions. The proposed architecture uses a centralized scheme that combines state-of-the-art, ad-hoc implementations of algorithms and techniques to maximize robustness and flexibility. It is based on a Model Predictive Controller (MPC) for which efficient cost function and constraint sets are designed to ensure a safe and accurate docking. A simulation environment is also presented to validate and test the proposed control scheme. 

\end{abstract}


\end{frontmatter}

\captionsetup{width=.45\textwidth}

\section{Introduction} \label{ch:chap_1}
\subsection{Motivation}
The subject of autonomous rendezvous and docking in space has become popular during the last decades. 
Successful proximity operations can be utilized to extend the life span of satellites and help with deorbiting when they reach end of life (EOL). The need of accurate pose estimation of uncooperative targets in space for navigation and docking is highlighted by the European Space Agency (ESA) hosting their Pose Estimation Challenge in 2019 \citep{esa_challenge}. This challenge also enforced the use of only a single camera, highlighting the importance of simple, compact and cheap sensors for this kind of applications. Most of the published research on this topic focus on single-agent docking missions. With the recent advancements on the launch of small platforms and cansats, the use of multiple, smaller and simpler satellites can greatly increase the efficiency of the mission in terms of time and cost, which is the main motivation of the present article.

\subsection{Related Works}
A family of solutions for the ESA challenge used Convolutional Neural Networks (CNN) to estimate the pose of the target. However, the most popular implementations either require an a priori knowledge of the targets wireframe e.g. \citep{cnn1}.
Some solutions have also been presented that lift this limitation e.g. \citep{cnn2}. 
Nevertheless, the use of CNNs for pose estimation is still a relatively new field and require huge amounts of data for training.

More conventional solutions for the problem have been proposed in the literature. \cite{image_based_control_Zhao} splits these solutions into two categories: image-based and position-based. An example of an image-based solution, a controller built directly on the image taken by the sensor, is proposed by \cite{Huang2017}.
This work focuses on detecting well defined edges on spacecraft parts, such as brackets, where a mechanical grabber can easily attach.
\cite{pose_estimation_depthcam_Harvey} propose matching a 3D model of the target with the output of on-board sensors but this method requires more complex LiDAR sensors.
An effective and accurate way to estimate the pose of the target is to use markers on its surface.
\cite{marker_docking_nasa} evaluated the use of special, reflective markers for docking purposes but still use complex combinations of LiDAR and camera sensors.

An example of a position-based approach is proposed by \cite{optical_aided_rendezvous_Renato} and is one of the most promising solutions in literature.
It uses a combination of a pose estimation method based on the Structure from Motion (SfM) algorithm and an optimal control scheme, removing the need for markers and using only a monocular camera.
But still, this work focuses on a single-agent docking scenario.
Examples of docking multiple satellites exist in the literature but mostly focus on them docking with each other and not with an uncooperative target e.g. \citep{ardc} and \citep{multi_6dog_gnc}.

\subsection{Contributions}


The proposed framework in this work aims to integrate and present improvements based on aforementioned state-of-art works. Thus, the main contributions in this work are as follows:
\begin{enumerate}
    \item We propose a novel centralized Nonlinear Model Predictive Controller architecture for multiple chaser spacecraft towards collaborative tracking and docking with an uncooperative, tumbling target spacecraft.
    \item A robust, fault-tolerant, vision-based pose estimation framework
    \item A comprehensive evaluations of the framework in a realistic simulation environment is carried out to prove the efficacy of the proposed work.
\end{enumerate}



\section{Problem Formulation}\label{ch:chap2}
The target is defined as an uncooperative object tumbling freely in 6-\acrshort{dof}, with constant velocities and with very simple markers imprinted on its body.
The environment is considered as a frictionless, gravity-free, non-interacting, infinite space with sufficient lighting. A chaser is assumed to be a robotic system, capable of 6-\acrshort{dof} motion in the environment and equipped with a monocular camera. The odometry of the \gls{chaser} is assumed to be known at any given time in relation to a static world frame.

The problem scenario discussed in this article is divided in to two parts. Initially, the synchronization problem is considered where the satellites will track the target and try to match it's tumbling path by maintaining a predefined pose $ref$, relative to the target. Later, satellites will slowly approach the target while maintaining their relative orientation until they are close enough to be considered docked to the target.

\subsection{Reference Frames}

\begin{figure}[h]
    \includegraphics[width=.45\textwidth]{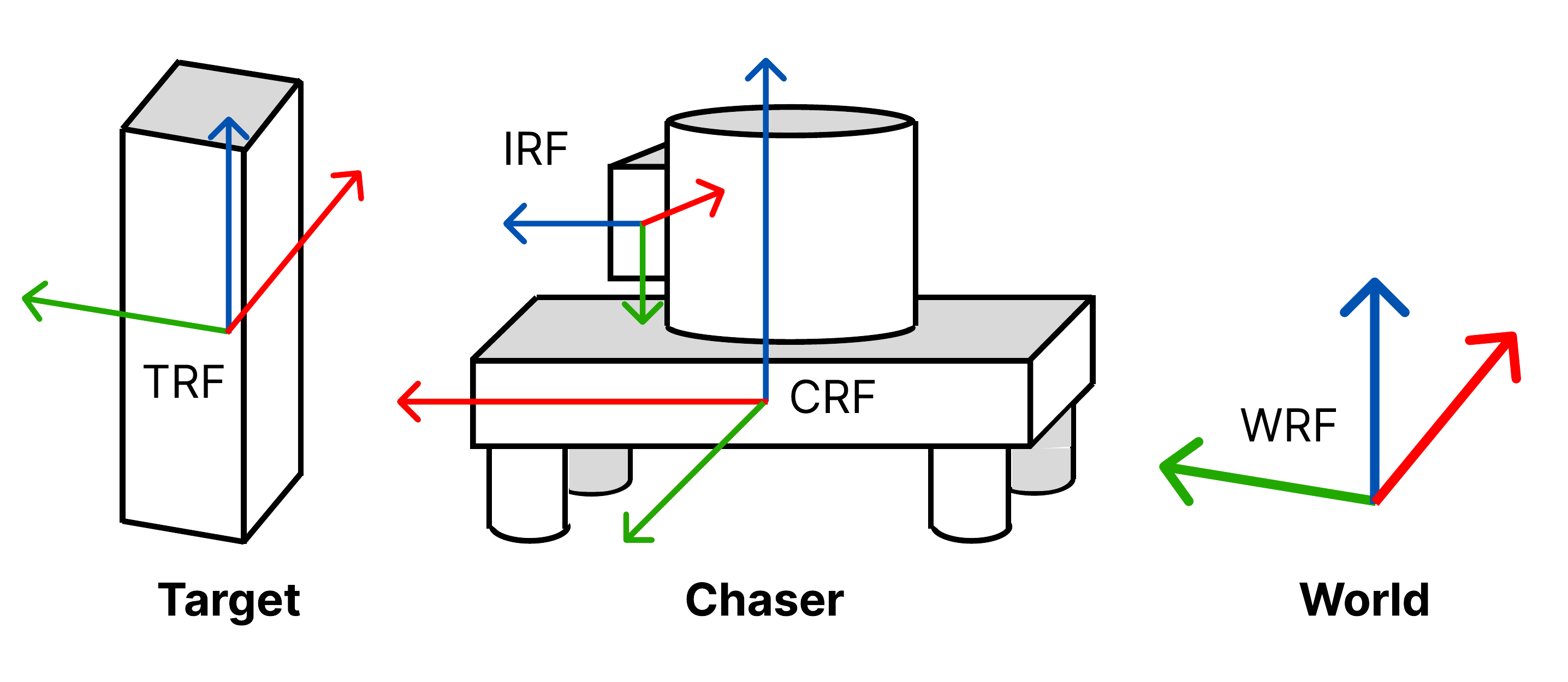}
    \centering
    \caption{The reference frames used in this work.}
    \label{fig:1_ref_frames}
\end{figure}

The reference frames used are presented in Fig. (\ref{fig:1_ref_frames}).

\begin{description}
    \item[WRF] the \textit{world frame}, is an arbitrary reference frame which will be used as a global reference.
    \item[CRF] the \textit{chaser frame} is a frame aligned with the \gls{chaser}'s chassis and fixed on its center of mass.
    The \textbf{X}-axis of ${CRF}$ aligns with the direction that the chaser's camera is looking towards.
    The \textbf{Z}-axis points towards the top of the chaser's chassis and \textbf{Y}-axis completes the right handed coordinate frame.
    \item[TRF] the \textit{target frame} is a frame fixed at the center of mass of the target and arbitrarily aligned.
    \item[IRM] the \textit{image frame}, is a necessary transformation to properly align the reference frames with the notation used by \gls{opencv} 
\end{description}

\subsection{System Dynamics} \label{ch:chap3:sec2}

It is assumed that both the target and chasers are rigid bodies. In terms of inertial properties, target and chasers are considered to be rectangular cuboids as well as there are no gravity or any external forces are acting on the system.

\subsubsection{Chaser}
The chaser can be described as a rigid body with forces and/or torques acting on it's center of gravity representing thruster actuation.
The chaser's state is defined by 13 variables:
$p=[p_x, p_y, p_z]^\top$ is the position, $q=[q_w, q_x, q_y, q_z]^\top \in [-1, 1]^4$ is a rotation quaternion representing the orientation and $v=[\dot{p_x}, \dot{p_y}, \dot{p_z}]^\top$ and $\omega=[\omega_x, \omega_y, \omega_z]^\top$ are the linear and angular velocities in Euler form, respectively.

It is assumed that all the forces and torques are acting on its center of gravity and are relative to the $CRF$ reference system.
To model the physics of the chaser the Euler-Newton equations for translational and rotational dynamics of a rigid body \eqref{eq:3_euler_newton} are used.
\begin{equation}\label{eq:3_euler_newton}
    \begin{bmatrix} F \\ \tau \end{bmatrix} = \begin{bmatrix} mI_3 & 0 \\ 0 & I_{cm} \end{bmatrix}
    \begin{bmatrix} \dot{v}(t) \\ \dot{\omega}(t) \end{bmatrix} + 
    \begin{bmatrix} 0 \\ \omega(t) \times (I_{cm} \cdot \omega(t))\end{bmatrix}
\end{equation}
where $F$ and $\tau$ are the forces and torques acting on the bodys center of gravity, $m$ and $I_{cm}$ are the mass and the moment of inertia matrix of the body. 

Equation.~\eqref{eq:3_chaser_dynamics} presents the complete linear and angular dynamics of the chaser spacecraft.
\begin{equation}\label{eq:3_chaser_dynamics}
    \left\{ \begin{matrix*}[l]
        \dot{p}(t) = v(t)\\
        \dot{v}(t) = \frac{1}{m}I_3\bm{\hat{F}}\\
        \dot{q}(t) = -\frac{1}{2} \begin{bmatrix} 0 \\ \omega(t) \end{bmatrix} \otimes q(t)\\
        \dot{\omega}(t) = I^{-1}_{cm} \cdot \bm{\hat{\tau}} - I^{-1}_{cm}[ \omega(t) \times (I_{cm} \cdot \omega(t)) ]
    \end{matrix*} \right.
\end{equation}
$\bm{\hat{\tau}}$ and $\bm{\hat{F}}$ represent the forces and torques acting on the center of mass of the chaser rotated back to the $WRF$ reference system.
This can be performed by using the following equations \eqref{forcerotate} of quaternions rotating vectors:
\begin{equation} \label{forcerotate}
    \begin{matrix}
        \bm{\hat{F}}=q\bm{F}q^*\\
        \bm{\hat{\tau}}=q\bm{\tau}q^*
    \end{matrix}
\end{equation}

\subsubsection{Target}
The target is also a rigid-body but with constant-velocity and no forces acting on it.
The target has similar state variables to the chaser's, i.e. $[p^{tar},q^{tar},v^{tar},\omega^{tar}]^\top$  while the system dynamics can be described by \eqref{eq:3_target_dynamics}.
\begin{equation}\label{eq:3_target_dynamics}
    \left\{ \begin{matrix*}[l]
        \dot{p}^{tar}(t) = v^{tar}(t)\\
        \dot{v}^{tar}(t) = 0\\
        \dot{q}^{tar}(t) = -\frac{1}{2} \begin{bmatrix} 0 \\ \omega^{tar}(t) \end{bmatrix} \otimes q^{tar}(t)\\
        \dot{\omega}^{tar}(t) = 0
    \end{matrix*} \right.
\end{equation}

\section{Proposed Methodology}\label{ch:chap2.5}

To tackle the scenario at hand, a centralized control scheme was designed as shown in Fig. (\ref{fig:1_chaser_flow}).
Each \gls{chaser} will process the images taken with its camera sensor on-board.
The images are un-distorted using a camera calibration model and then passed to the pose estimator.

\insertfig{1_chaser_flow}{Processing flow of a chaser swarm composed of two agents.}

The pose estimator starts by detecting the ArUco markers \citep{aruco_1}, \citep{aruco_2} imprinted on the target.
The positions of the marker's corners in the $TRF$ frame define a set $P_{TRF}$.
The method for ArUco marker recognition allows extraction of the pixel coordinates of each individual corner of each marker.
Each marker is also uniquely identifiable.
A set $P_{IRF}$ is defined containing these 2D coordinates of the corners visible on a given image capture.

Next, a \acrlong{pnp} (PnP) problem is defined using $P_{TRF}$ and $P_{IRF}$.
The \acrshort{pnp} problem is the problem of determining the pose of a calibrated camera given a set of $n$ 3D points in the world and their corresponding 2D projections in the image \citep{springer_robotics:ch32, pnp_1}.
Basically, solving the pinhole camera model in (\ref{eq:pinhole}) for the extrinsic parameters ($[R|t]$), provided some 3D world coordinates ($M$) and 2D image coordinates ($m$) and with the intrinsic parameters ($A$) known.

\begin{subequations} \label{eq:pinhole}
    \begin{equation}
        s m' = A [R|t] M'
    \end{equation}
    \begin{equation}
        s \begin{bmatrix} u \\ v \\ 1 \end{bmatrix} = 
        \begin{bmatrix}
            f_x & 0 & c_x \\
            0 & f_y & c_y \\
            0 & 0 & 1
        \end{bmatrix} 
        \left[
        \begin{array}{ccc|c}
            R_{11} & R_{12} & R_{13} & T_1\\
            R_{21} & R_{22} & R_{23} & T_2\\
            R_{31} & R_{32} & R_{33} & T_3
        \end{array}
        \right]
        \begin{bmatrix} X \\ Y \\ Z \\ 1 \end{bmatrix}
    \end{equation}
\end{subequations}

Implementations for both ArUco marker detection algorithm and the EPnP \citep{epnp} solver used in this project were provided in the OpenCV library.

The target's pose estimations are sent to a TF system responsible for calculating transfer functions between different reference frames.
An estimation combiner node is responsible to process the pose estimates of the \gls{chaser}s and to combine them into a single estimate for the \gls{target}s odometry (pose \& velocities).
The combiner works by averaging the latest available estimations by each chaser that has line-of-sight to the target.
It also uses a moving average filter for noise reduction, outlier detection and removal methods.

This combined estimation is fed, together with the chaser's odometry, to the controller.
The controller then uses the estimated odometry to generate an optimal control signal for each chaser (see Ch. \ref{ch:chap3}).
The control signals, consisting of a set of 3-dimensional forces and torques, are sent to the \gls{chaser}'s Thruster Selection Logic (TSL) and finally to a Pulse Width Modulation (PWM) signal generator.
The TSL module is considered as a black box for now and will be implemented later ad-hoc.

\section{Controller} \label{ch:chap3}

As previously mentioned, the control architecture uses a MPC backbone.
In this section, the cost function and constraints used to formulate the MPC will be presented.





\subsection{Cost Function} \label{ch:chap3:sec3}



The chasers state vectors are defined as $x^i = [p^i, q^i]^\top$ and the corresponding control action as $u^i = [F^i, \tau^i]^\top$.
The system dynamics are discretized with a sampling time of $dt$ using the forward Euler method to obtain $x^i_{k+1} = \zeta(x^i_k, u^i_k)$. 
The target's state vector is defined as $x^{tar} = [p^{tar}, q^{tar}]$, where $p^{tar}$ is the position and $q^{tar}$ the quaternion attitude represention of the target.
They are descritized in a manner similar to the chaser's states.
The reference poses for the chasers are calculated by transfering the target's body frame using $x^{{off}_i} = [p^{{off}_i}, q^{{off}_i}]$, where $p^{{off}_i}$ is a translation and $q^{{off}_i}$ is a rotation quaternion.
These calculations are performed as such: $p^{{ref}_i}_k = p^{tar}_k + q^{tar}_k \otimes p^{{off}_i} \otimes {q^{tar}_k}^\ast$ and $q^{{ref}_i}_k = q^{tar}_k q^{{off}_i}$.
This discrete model is used as the prediction model of the NMPC.
The prediction is performed over a receding horizon of $D = T/dt$ steps, where $T$ is the horizon duration in seconds.

A cost function is defined such that, when minimized in the current time and the predicted horizon, an optimal set of control actions $u^i_k$ will be calculated.
Let $x_{k+j|k}$ and $x^{tar}_{k+j|k}$ be the predicted chaser and target states at time step $k+j$ respectively, calculated in time step $k$.
The corresponding control actions are $u_{k+j|k}$ and reference states $x^{{ref}_i}_{k+j|k}$.
Also, let $\bm{x}_k$ and $\bm{u}_k$ be the predicted states and control actions for the whole horizon duration calculated at time step $k$.
The cost function is formulated as follows:

\begin{multline} \label{eq:3_cost_function}
    J(\bm{x}_k, \bm{u}_k, u_{k-1|k}) = \sum_{i=0}^{N-1} \{ \sum_{j=0}^{D} \{ \\ 
    \underbrace{(1 - \frac{\alpha \cdot k}{D} )}_\text{Falloff \%} \cdot [\underbrace{{\| p^{{ref}_i}_{k+j|k} - p^i_{k+j|k} \|}^2 Q_p}_\text{Position cost} + \underbrace{{\| q^i_{k+j|k} \otimes {q^{{ref}_i}_{k+j|k}}^\ast \|}^2 Q_q}_\text{Orientation cost}] \\
    + \underbrace{\| u^i_{ref} - u^i_{k+j|k} \|^2 Q_u \}}_\text{Input cost} \\ 
    + \underbrace{{\| p^{{ref}_i}_{k+D|k} - p^i_{k+D|k} \|}^2 Q^f_p + {\| q^i_{k+D|k} \otimes {q^{{ref}_i}_{k+D|k}}^\ast \|}^2 Q^f_q}_\text{Final state cost} \}
\end{multline}

where $Q_p, Q^f_p \in \mathbb{R}^{3\times3}$, $Q_q, Q^f_q \in \mathbb{R}^{4\times4}$ and $Q_u \in \mathbb{R}^{6\times6}$ are positive definite weight matrices for the position and orientation states and inputs respectively. 
In \eqref{eq:3_cost_function}, the first term represents the state cost which penalizes deviation from the reference state at time-step $k$, $x^{{ref}_i}_k$.
The Falloff term $\alpha$, is an adaptive weight to penalize overshoot errors.
The second term represents the input cost which penalizes deviation from the steady-state input $u_{ref} = 0$ that describes constant-velocity movement.
Finally, the final state-cost applies an extra penalty for deviation of the state from the reference at the end of the horizon period.


\subsection{Constraints} \label{ch:chap3:sec4}

A minimum chaser-chaser and chaser-target distance $d_{min}$ is enforced by \eqref{eq:3_constr_dist1} and \eqref{eq:3_constr_dist2}.
\eqref{eq:3_constr_dist3} prevents the chasers from moving too far away from the target by setting their maximum distance to $d_{max}$.
The implemented constraints are the following:
\begin{subequations}
    \begin{equation}\label{eq:3_constr_max}
        u^i_{k+j|k} = [\bm{F}^i_{k+j|k}, \bm{\tau}^i_{k+j|k}] \in [-F_{max}, F_{max}] \times [-\tau_{max}, \tau_{max}]
    \end{equation}
    
    \begin{equation}\label{eq:3_constr_dist1}
        \dist{(p^{tar}_{k+j|k}, p^m_{k+j|k})} \geq d_{min}\; \forall\; j \in [0, D]
    \end{equation}
    
    \begin{equation}\label{eq:3_constr_dist2}
        \dist{(p^m_{k+j|k}, p^n_{k+j|k})} \geq d_{min} \forall\; j \in [0, D],\; m \neq n
    \end{equation}
    
    \begin{equation}\label{eq:3_constr_dist3}
        \dist{(p^{{ref}_m}_{k+j|k}, p^m_{k+j|k})} \leq d_{max}\; \forall\; j \in [0, D]
    \end{equation}
\end{subequations}
where $dist$ is a function of Eucledean distance.

\subsection{Optimizer} \label{ch:chap3:sec5}



\acrfull{open} was used as the MPC cost function optimizer.
\acrshort{open} is a framework developed by \cite{opengen} and is based on the PANOC \citep{panoc} optimization algorithm.
PANOC is an extremely fast, the state-of-the-art optimizer for real-time, embedded applications.
A powerful symbolic mathematics library called CasADi \citep{casadi} is used to define the optimization problem and perform under-the-hood operations.


\section{Simulation Results} \label{ch:chap4}
The following section presents the results of the simulations that were performed to test the control architecture.
First of all, in Section \ref{ch:chap4:sec1} different parameters of the MPC controller are tested and compared in a single chaser scenario.
Then, in Section \ref{ch:chap4:sec2} multi chaser scenarios are performed and the docking sequence is tested.
Finally, in Section \ref{ch:chap4:sec3} a clear demonstration of the collision avoidance capabilities are presented.
In every scenario the velocities of the target are $v=[1.50, 0.75, 3.00]\times10^{-2}\text{m/s}$ and $\omega=[1.50, 4.50, 3.00]\times10^{-2}\text{rad/s}$ and the simulations were recorded over 2 minutes.
A video demonstration of these simulations can be found at \textcolor{blue}{youtu.be/86JT4M7eyGc}.

\subsection{Single Chaser \& MPC Tuning} \label{ch:chap4:sec1}

The tuneable parameters of a MPC controller are the cost function parameters described in \eqref{eq:3_cost_function}.
The error terms of each state variable, scaled by their weight $Q_p$, need to be close to each other to avoid some to be prioritized over others.
For example in cases where the chasers start far away from the target, the position error will be degrees of magnitude higher than the quaternion error which is bounded in $[-1, 1]$.
The $Q_u$ weights penalize the use of the control inputs and can be tuned by trail and error to avoid over-actuation of the chaser.
Finally the $Q_f$ weights penalize the final state error to avoid overshooting the target and must be tuned higher than $Q_p$.

\begin{table}[h]
    \centering
    \begin{tabular}{|c|c|c|c|}
        \hline
         & $Q_p$ & $Q_u$ & $Q_f$ \\
        \hline
         Position/Force & $65$ & $3.5$ & $3250$\\
        \hline
         Orientation/Torque & $35$ & $40$ & $1750$\\
        \hline
    \end{tabular}
    \vspace*{.15cm}
    \caption{MPC Tuning Weights}
    \label{tab:weights}
\end{table}
\vspace*{-.25cm}

\begin{figure}
    \centering
    \resizebox{.49\textwidth}{!}{
        \input{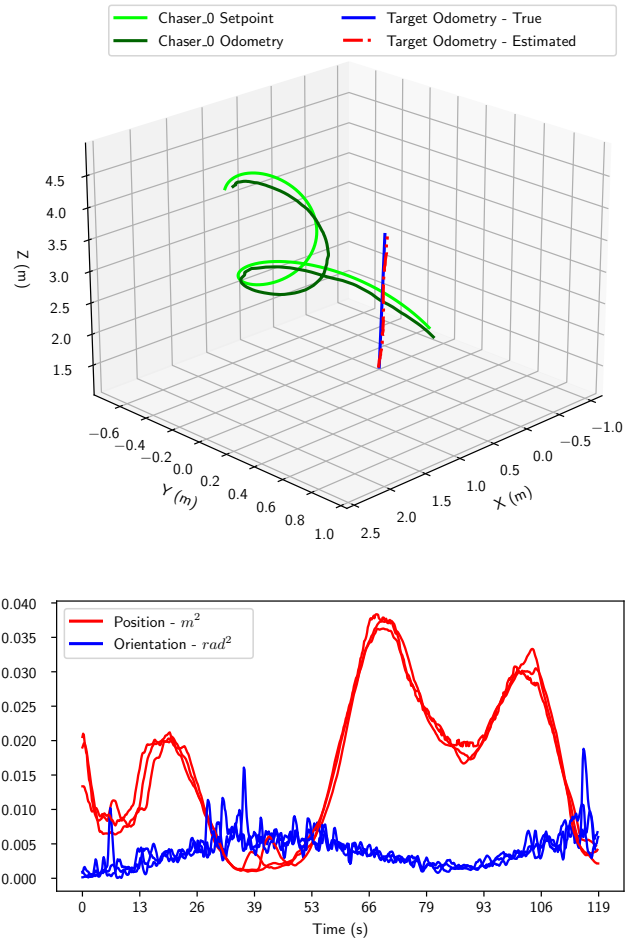}
    }
    \caption{Baseline runs with 1 chaser, $T=3\text{s}$, $dt=0.1\text{s}$ and $\alpha=0\%$. \textbf{Top:} Chaser and target trajectories, \textbf{Bottom:} Pose errors.}
    \label{fig:baseline_1p}
\end{figure}

The values of these weights are presented in Table \ref{tab:weights} and were determined after repetitive testing.
The rest of the cost function parameters have more interesting effects on the performance of the controller and thus are further analyzed in the following subsections.
As a baseline for comparison, Figure \ref{fig:baseline_1p} shows the results of the MPC controller with $T=3\text{s}$, $dt=0.1\text{s}$ and $\alpha=0\%$.
Every test was performed and recorded three times for more accurate results.
Table \ref{tab:1p_tests}  presents the results of the simulations.

\begin{table}[h]
    \centering
    \begin{tabular}{|m{1.5cm}|c|c|c|c|c|}
        \hline
        $\bm{Comment}$ & $\bm{T}$(s) & $\bm{dt}$(s) & $\bm{\alpha}$(\%) & \thead{Position\footnotemark \\ $\bm{MSE}$} & \thead{Orientation \\ $\bm{MSE}$}\\ 
        \hline
        Baseline & $3$ & $1/10$ & $0\%$ & $1.64$ & $3.94$ \\
        \hline
        \multirow{2}{*}{$\bm{T}$ test}
            & $0.5$  & $1/10$ & $0\%$ & $64.8$\footnotemark & $229$ \\
            \cline{2-6}
            & $10$  & $1/10$ & $0\%$ & $126$\footnotemark & $943$ \\
        \hline
        \multirow{2}{*}{$\bm{dt}$ test}
            & $3$  & $1/5$ & $0\%$ & $1.73$ & $4.87$ \\
            \cline{2-6}
            & $3$  & $1/30$ & $0\%$ & $1.42$ & $3.89$ \\
        \hline
        \multirow{3}{*}{$\bm{\alpha}$ test}
            & $3$  & $1/10$ & $10\%$ & $1.65$ & $4.09$\\
            \cline{2-6}
            & $3$  & $1/10$ & $30\%$ & $1.69$ & $4.07$ \\
            \cline{2-6}
            & $3$  & $1/10$ & $60\%$ & $1.75$ & $4.25$ \\
            \cline{2-6}
            & $3$  & $1/30$ & $30\%$ & $1.59$ & $4.02$ \\
        \hline
    \end{tabular}
    \vspace*{.15cm}
    \caption{Single Chaser Tests. Position MSE: $m^2 \times 10^{-2}$, Orientation MSE: ${rad}^2 \times 10^{-3}$}
    \label{tab:1p_tests}
\end{table}
\vspace*{-.25cm}
\footnotetext[2]{Average over 3 runs.}
\footnotetext[3]{1/3 runs failed.}
\footnotetext[4]{2/3 runs failed.}

\subsubsection{Horizon Period:} \label{ch:chap4:sec1:sub1}
$T$ is the time in seconds that the controller will predict the future of the system.
It is clear that $T$ is the most important parameter for tuning the MPC controller.
Having too small a horizon allows the controller to predict only the short term results of the plant's input and too large a horizon will make the controller too slow.
Both tests with $T=0.5\text{s}$ and $T=10\text{s}$ had failed runs.

\subsubsection{Prediction Time Step:} \label{ch:chap4:sec1:sub2}
$dt$ is the sampling for the horizon period. 
Smaller $dt$ values will make the controller more accurate since it improves the performance of the forward Euler method. 
Decreasing $dt$ also adds more steps to the MPC and thus more variables in the optimization problem, increasing the calculation time for a solution.

\subsubsection{Falloff percentage:} \label{ch:chap4:sec1:sub3}
The experimental term $\alpha$ is tested for performance improvement.
The test results show that $\alpha$ actually increases the MSE of every scenario and thus is not used in the final controller.


\subsection{Multi Chaser} \label{ch:chap4:sec2}

\begin{figure}
    \centering
    \resizebox{.49\textwidth}{!}{
        \input{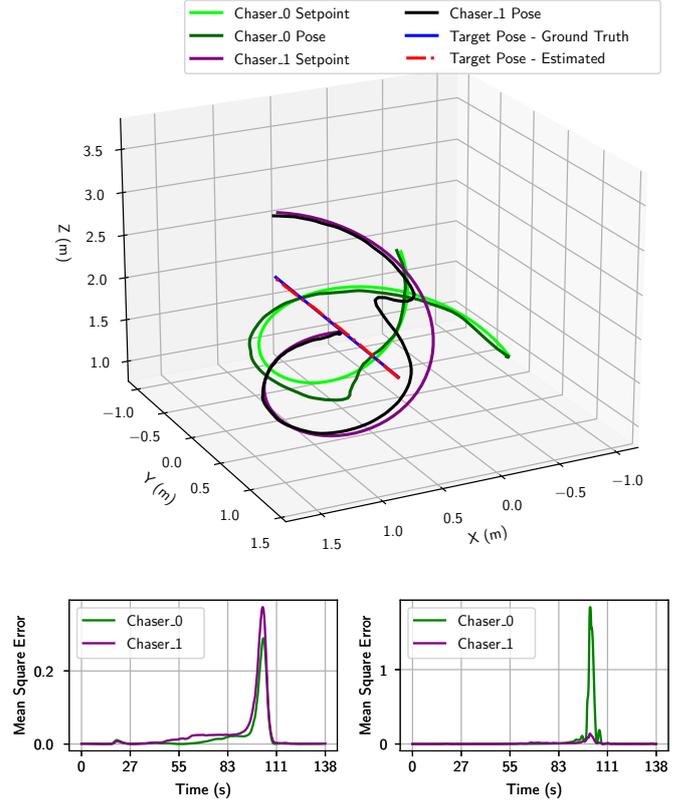}
    }
    \caption{Two Chaser tracking mission. \textbf{Top:} Chaser and target trajectories, \textbf{Bottom left:} Chaser position error, \textbf{Bottom right:} Chaser orientation error}
    \label{fig:2p_test}
\end{figure}

For multi chaser scenarios, two tests were performed.
The first test is a simple tracking scenario where two chasers communicate their estimations of the target's pose to each other to maintain a pre-defined pose relative to the target.
In Figure \ref{fig:2p_test}, a spike in the MSE plot is observed at around 100s, which is caused by a Euler angle singularity introduced by the PnP solver of OpenCV.
This singularity can be interpreted as a random interference in the system which the controller was able to handle, hence showcasing it's robustness.

\begin{figure}
    \centering
    \resizebox{.49\textwidth}{!}{
        \input{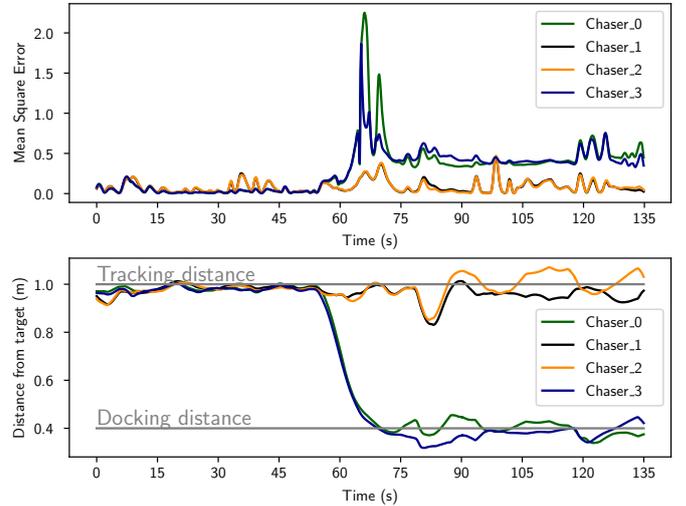}
    }
    \caption{Four Chaser tracking mission with Chaser 0 and 3 docking after 60 seconds. \textbf{Top: } Chasers pose error, \textbf{Bottom: } Docking evaluation}
    \label{fig:4p_test}
\end{figure}

The second test is a more complex scenario where four chasers initially track the target.
After about 60s Chaser 0 and 3 receive an approach command and start docking the target.
Their relative pose is maintained throughout the docking process but their position is gradually scaled to decrease their distance to the target.
Figure \ref{fig:4p_test} showcases the distance each chaser maintains from the target.
Of course the pose error increases in the docking process since the chaser's position is compared to their initial $ref$ pose and not the docking pose.

\subsection{Collision Avoidance} \label{ch:chap4:sec3}

To showcase the collision avoidance capabilities of the controller, a simple scenario was created where two chasers are initially just tracking a stationary target.
A command is given to Chaser 0 to change it's $ref$ pose to a position on the opposite side of the target.
The closest route to that pose would be a direct line through the target.
However, the controller is able to find a path over the target, respecting a limit of $d_{min}=0.35$m.
Chaser 1 maintains it's initial $ref$ pose relative to the static target, observing the scene from afar.
This is crucial since Chaser 0 will inevitably loose sight of the target and be unable to estimate it's pose alone.
The pose estimation from Chaser 1 is used by Chaser 0 during the maneuver, showcasing one of the benefits of the multi chaser approach.

\begin{figure}
    \centering
    \resizebox{.49\textwidth}{!}{
        \input{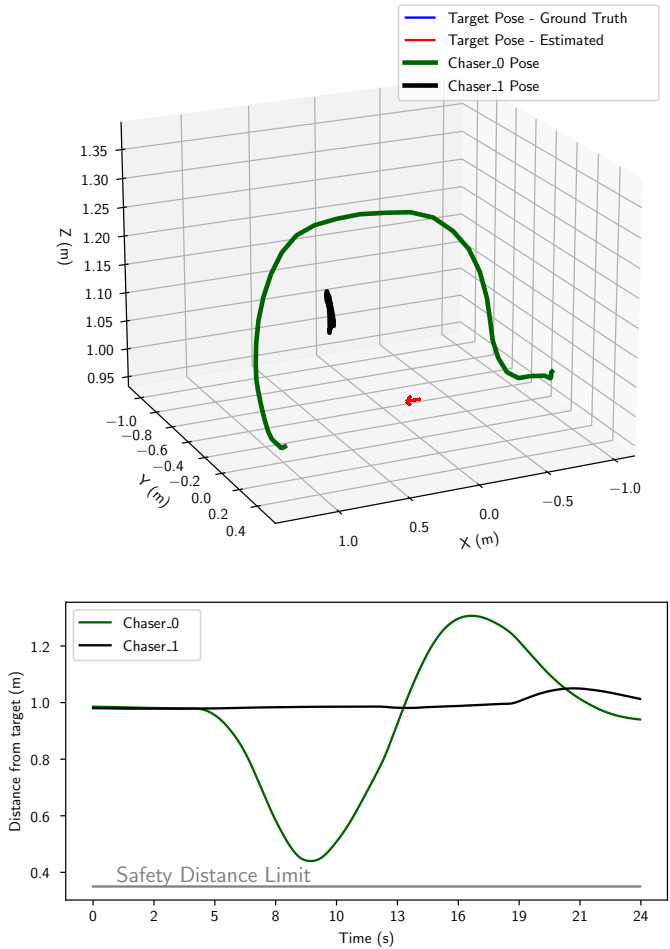}
    }
    \caption{Two Chaser, collision avoidance mission. Chaser 0 maneuvers over the target while Chaser 1 assists with visual pose estimation. \textbf{Top: } Chaser trajectories and Target pose,
    \textbf{Bottom: } Collision avoidance evaluation}
    \label{fig:collision_avoidance}
\end{figure}


\section{Conclusion}
In this article we propose a reliable centralized NMPC-based control architecture that can assist in docking of multiple chaser spacecraft to a tumbling target in Space. The tumbling target pose is estimated using a single monocular camera on-board each chaser spacecraft. The estimated pose is shared between the chaser spacecraft to improve the fused pose estimate of the target when the target moves of the field of view of one of the chaser spacecraft. Multiple simulation scenarios were tested to determine the efficacy of the proposed controller.  


\bibliography{ifacconf}

\end{document}